\documentclass[11pt,a4paper]{article}
\usepackage[hyperref]{acl2021}
\usepackage{times}
\usepackage{latexsym}

\usepackage{microtype}

\usepackage{bm}
\usepackage{amsmath}
\usepackage{amssymb}
\usepackage{graphicx}
\usepackage{url}
\usepackage{multirow}
\usepackage{color}
\usepackage{makecell}
\usepackage{subfigure}
\usepackage{bbding}

\aclfinalcopy 
\def\aclpaperid{1686} 

\title{From Discourse to Narrative: Knowledge Projection \\for Event Relation Extraction}

\author{
\textbf{Jialong Tang}${}^{1,3}$, \textbf{Hongyu Lin}${}^{1}$, \textbf{Meng Liao}${}^{4}$, \textbf{Yaojie Lu}${}^{1,3}$, \\
\textbf{Xianpei Han}${}^{1,2}$, \textbf{Le Sun}${}^{1,2,*}$, \textbf{Weijian Xie}${}^{4}$, \textbf{Jin Xu}${}^{4,*}$\\
${}^{1}$Chinese Information Processing Laboratory ~ ${}^{2}$State Key Laboratory of Computer Science\\
Institute of Software, Chinese Academy of Sciences, Beijing, China\\
${}^{3}$University of Chinese Academy of Sciences, Beijing, China\\
${}^{4}$Data Quality Team, WeChat, Tencent Inc., China\\
\texttt{\{jialong2019,hongyu,xianpei,sunle\}@iscas.ac.cn} \\
\texttt{\{maricoliao, vikoxie, jinxxu\}@tencent.com} \\
}

\date{}

\begin{document}
\maketitle
\begin{abstract}
{
  \renewcommand{\thefootnote}{\fnsymbol{footnote}}
}
Current event-centric knowledge graphs highly rely on explicit connectives to mine relations between events. 
Unfortunately, due to the sparsity of connectives, these methods severely undermine the coverage of EventKGs. 
The lack of high-quality labelled corpora further exacerbates that problem. 
In this paper, we propose a knowledge projection paradigm for event relation extraction: projecting discourse knowledge to narratives by exploiting the commonalities between them. 
Specifically, we propose \textbf{M}ulti-tier \textbf{K}nowledge \textbf{P}rojection \textbf{Net}work (\textbf{MKPNet}), which can leverage multi-tier discourse knowledge effectively for event relation extraction.
In this way, the labelled data requirement is significantly reduced, and implicit event relations can be effectively extracted. 
Intrinsic experimental results show that MKPNet achieves the new state-of-the-art performance, and extrinsic experimental results verify the value of the extracted event relations.
\let\thefootnote\relax\footnotetext{${}^{*}$Corresponding authors.}
\end{abstract}

\section{Introduction} 
\label{sec:introduction}

\begin{figure}[!t]
\centering
\includegraphics[width=0.95\columnwidth]{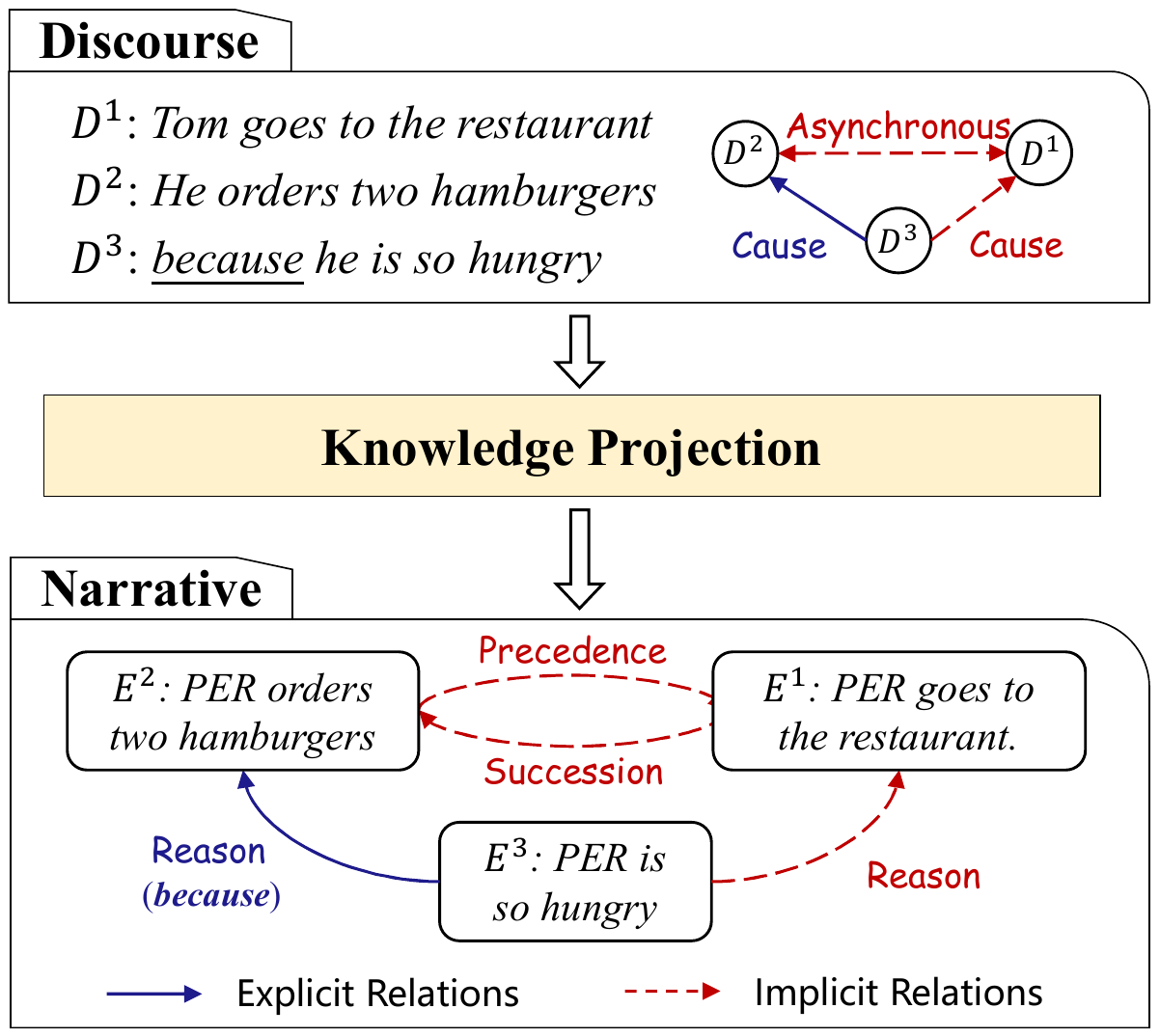}
\caption{
The knowledge projection paradigm for event relation extraction. 
The explicit projection directly projects connectives to event relations, e.g., from \emph{``because''} to \emph{Reason}. 
The implicit projection leverages the discourse knowledge to discover implicit event relations without connectives via MKPNet.}
\label{fig:introduction}
\end{figure}

Event-centric knowledge graphs (EventKGs) model the narratives of the world by representing events and identifying relations between them, which are critical for machine understanding and can benefit many downstream tasks, such as question answering~\citep{costa2020event}, news reading~\citep{vossen2018newsreader}, commonsense knowledge acquisition~\citep{zhang2020transomcs} and so on.

Recently, semi-automatically constructing EventKGs have gained much attention~\citep{tandon2015knowlywood,rospocher2016building,gottschalk2018eventkg,zhang2020aser}. 
These methods extract event knowledge from massive raw corpora with or without little human intervention, which makes them scalable solutions to build large-scale EventKGs. 
Commonly, each node in EventKGs represents an event, and each edge represents a pre-defined relation between an event pair\footnote{
Computational and cognitive studies define nodes as eventualities, which include activities, states and events.
In this paper, we simplify the definition of each node to ``event'' due to its popularity.
}.
Currently, event relations are majorly extracted based on the explicit connectives between them.
For example, in Figure~\ref{fig:introduction}, a \emph{Reason} relation is extracted between $E^2$: \emph{``PER orders two hamburgers''} and $E^3$: \emph{``PER is so hungry''} using the explicit connective \emph{``because''} between them.

Unfortunately, the connective-based approaches face the critical coverage problem due to the sparsity of connectives.
That is, a large proportion of event pairs are not connected with explicit connectives, but with underlying event relations.
We denote them as \textbf{\emph{implicit event relations}}. 
Furthermore, the related events can even not close to each other in a document.
For the example in Figure~\ref{fig:introduction}, the implicit relation \emph{Reason} between $E^1$: \emph{``PER goes to the restaurant''} and $E^3$: \emph{``PER is so hungry''} can not be extracted due to the absence of explicit connective as well as the discontinuity between these two clauses.
The common practice in previous connective-based approaches is to ignore all these implicit instances~\citep{zhang2020aser}. 
As a result, the coverage of EventKGs is significantly undermined. 
Besides, because the scale of the existed event relation corpus~\citep{hong2016building} is limited, it is also impractical to build effective event relation classifiers via supervised learning.

In this paper, we propose a new paradigm for event relation extraction --- knowledge projection. 
Instead of relying on sparse connectives or building classifiers starting from scratch, we project discourse knowledge to event narratives by exploiting the anthropological linguistic connections between them. 
Enlightened by~\citet{livholts2015discourse,altshuler2016events,reyes2017discourse}, discourses and narratives have significant associations, and their knowledge are shared at different levels: 
1) token-level knowledge: discourses and narratives share similar lexical and syntactic structures, 
2) semantic-level knowledge: the semantics entailed in discourse pairs and event pairs are analogical, e.g., $E^3$-\emph{Reason}$\rightarrow$$E^1$ and $D^3$-\emph{Cause}$\rightarrow$$D^1$ in Figure~\ref{fig:introduction}., 
and 3) label-level knowledge: heterogeneous event and discourse relations have the same coarse categories, e.g., both the event relation \emph{Reason} and the discourse relation \emph{Cause} are included in the coarse-grained relation \emph{Contingency}. 
By exploiting the rich knowledge in manually labelled discourse corpus and projecting them into event relation extraction models, the performance of event relation extraction can be significantly improved, and the data requirement can be dramatically reduced.

Specifically, we design \textbf{M}ulti-tier \textbf{K}nowledge \textbf{P}rojection \textbf{Net}work (\textbf{MKPNet}), which can leverage multi-tier discourse knowledge effectively for event relation extraction. 
MKPNet introduces three kinds of adaptors to project knowledge from discourses into narratives: 
(a) token adaptor for token-level knowledge projection;
(b) semantic adaptor for semantic-level knowledge projection;
(c) coarse category adaptor for label-level knowledge projection.
By sharing the parameters of these three adaptors, the commonalities between discourses and narratives at various levels can be effectively explored.
Therefore, we can obtain more general token representations, more accurate semantic representations, and more credible coarse category representations to better predict event relations.

We conduct intrinsic experiments on ASER~\citep{zhang2020aser}, one of the representative EventKGs, and extrinsic experiments on Winograd Scheme Challenge (WSC)~\citep{levesque2012winograd}, one of the representative natural language understanding benchmarks.
Intrinsic experimental results show that the proposed MKPNet significantly outperforms the state-of-the-art (SoA) baselines, and extrinsic experimental results verify the value of the extracted event relations\footnote{
Our source codes with corresponding experimental datasets and the enhanced EventKG are openly available at \url{https://github.com/TangJiaLong/Knowledge-Projection-for-ERE}.
}.

The main contributions of this paper are:
\begin{itemize}
\item We propose a new knowledge projection paradigm, which can effectively leverage the commonalities between discourses and narratives for event relation extraction.
\item We design MKPNet, which can effectively leverage multi-tier discourse knowledge for event relation extraction via token adaptor, semantic adaptor and coarse category adaptor.
\item Our method achieves the new SotAevent relation extraction performance, and an enriched EventKG is released by extracting both explicit and implicit event relations. 
We believe it can benefit many downstream NLP tasks.
\end{itemize}

\begin{figure*}[!t]
\centering
\includegraphics[width=0.95\textwidth]{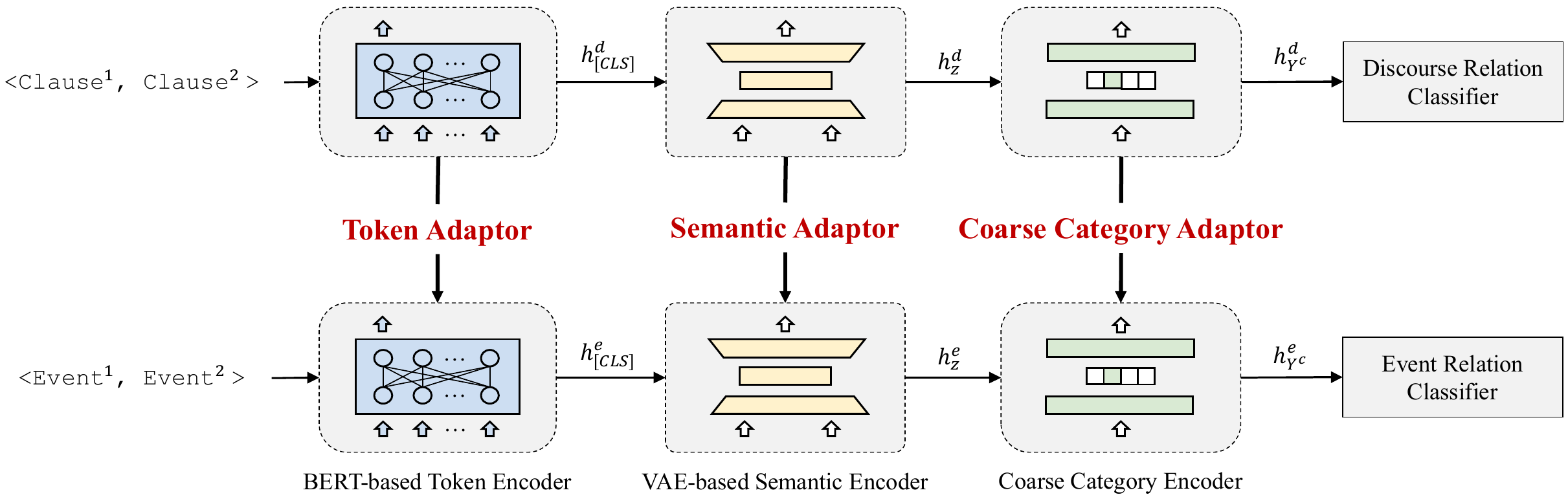}
\caption{An overview of MKPNet, which projects discourse knowledge for event relation extraction: (a) token adaptor for token-level knowledge projection, (b) semantic adaptor for semantic-level knowledge projection, and (c) coarse category adaptor for label-level knowledge projection.
}
\label{fig:mkpnet}
\end{figure*}

\section{Background} 
\label{sec:background}

\textbf{Event Relation Extraction (ERE).}
Given an existing EventKG $\mathcal{G}=\{\mathcal{E}, \mathcal{R}\}$, where nodes $\mathcal{E}$ are events and edges $\mathcal{R}$ are their relations.
$Y^{ex} \in \mathcal{R}$ are explicit event relations extracted by connective-based methods, and $Y^{im} \notin \mathcal{R}$ are implicit event relations without connectives.
Commonly, implicit event relation extraction (IERE) takes two events $E^1=\{e_1^1, ..., e_{|E^1|}^1\}$, $E^2=\{e_1^2, ..., e_{|E^2|}^2\}$, $E^1, E^2 \in \mathcal{E}$ as inputs, then uses a neural network to classify their underlying relation. 

\textbf{Discourse Relation Recognition (DRR).}
DRR aims to recognize the relation of two discourse arguments.
Discourse relations can be explicit or implicit, where explicit relations are revealed by connectives, while implicit relations lack these surface cues.
To resolve the implicit discourse relation recognition (IDRR) task, researchers construct high-quality labelled datasets~\citep{prasad2008penn} and design elaborate models~\citep{zhang2016variationala,bai2018deep,kishimoto2020adapting}.

\textbf{Associations between Discourse and Narrative.}
Recent NLP studies have proved that discourse and narratives closely interact with each other, and leveraging discourse knowledge benefits narrative analysis significantly, such as subevents detection~\citep{aldawsari2019detecting} and main event relevant identification~\citep{choubey2020discourse}.
Motivated by the above observation, this paper leverages the knowledge of discourse by a knowledge projection paradigm.
Blessed with the associations at token-, semantic- and coarse category-levels, the discourse corpora and knowledge can be effectively exploited for event relation extraction.

\section{Multi-tier Knowledge Projection Network for Event Relation Extraction} 
\label{sec:multi_tier_knowledge_projection_network_for_event_relation_extraction}

In this section, we describe how to learn an effective event relation extractor by projecting resource-rich discourse knowledge to the resource-poor narrative task. 
Specifically, we propose \textbf{M}ulti-tier \textbf{K}nowledge \textbf{P}rojection \textbf{Net}work (\textbf{MKPNet}) which can effectively leverage multi-tier discourse knowledge for implicit event relation extraction.
Figure~\ref{fig:mkpnet} shows an overview of MKPNet, which uses token adaptor, semantic adaptor and coarse category adaptor to fully exploit discourse knowledge at different levels.
In the following, we first describe the neural architecture of MKPNet and then describe the details of three adaptors.

\subsection{Neural Architecture of MKPNet} 
\label{subsec:neural_architecture_of_mkpnet}

For knowledge projection, we model both event relation extraction (ERE) and discourse relation recognition (DRR) as an instance-pair classification task~\citep{devlin2019bert,kishimoto2020adapting}.
For ERE, the input is an event pair such as \textless$E^1$: \emph{``PER goes to the restaurant''}, $E^3$: \emph{``PER is so hungry''}\textgreater~and the output is an event relation such as \emph{Reason}.
For DRR, the input is a clause pair such as \textless$D^1$: \emph{``Tom goes to the restaurant''}, $D^3$:\emph{``he is so hungry''}\textgreater~and the output is a discourse relation such as \emph{Cause}.

Specifically, MKPNet extends the SotADRR model --- BERT-CLS~\citep{kishimoto2020adapting} by the VAE-based semantic encoder and the coarse category encoder to model knowledge tier-by-tier~\citep{pan2016domain,guo2019spottune,kang2020decoupling,li2020bert}.
It 1) first utilizes the BERT-based token encoder to encodes an instance pair as a token representation $h_{[CLS]}$;
2) then obtains the semantic representation $h_z$ via a VAE-based semantic encoder;
3) predicts the coarse-grained label and embeddings it as the coarse category representation $h_{Y^c}$;
4) finally classifies its relation with the guidance of the aggregate instance-pair representation:
\begin{equation}
Y = Classifier^{Fine}([h_{[CLS]} \oplus h_z \oplus h_{Y^c}])
\end{equation}
where $\oplus$ means the concatenation operation.
In this way, the parameters of MKPNet can be grouped by $\{\theta^{BERT}, \theta^{Semantic}, \theta^{Coarse}, \theta^{Fine}\}$, where $\theta^{BERT}$ for BERT-based token encoder, $\theta^{Semantic}$ for VAE-based semantic encoder, $\theta^{Coarse}$ for coarse category encoder and $\theta^{Fine}$ for the final relation classifier layer respectively.

\subsection{Token Adaptor} 
\label{subsec:token_adaptor}

Recent studies have shown that similar tasks usually share similar lexical and syntactic structures and therefore lead to similar token representations~\cite{Pennington2014glove,peters2018deep}.
The token adaptor tries to improve the token encoding for ERE by sharing the parameters $\theta^{BERT}$ of the BERT-based encoders with DRR. 
In this way, the encoder is more effective due to the more supervision signals and is more general due to the multi-task settings.

Specifically, given an event pair \textless $E^1$, $E^2$\textgreater, we represent it as a sequence:
\begin{equation}\nonumber
[CLS], e_1^1, ..., e_{|E^1|}^1, [SEP], e_1^2, ..., e_{|E^2|}^2, [SEP]
\end{equation}
where[CLS] and [SEP] are special tokens. 
For each token in the input, its representation is constructed by concatenating the corresponding token, segment and position embeddings.
Then, the event pair representation will be inputted into BERT architecture~\citep{devlin2019bert} and updated by multi-layer Transformer blocks~\citep{vaswani2017attention}.
Finally, we obtain the hidden state corresponding to the special [CLS] token in the last layer as the token-level event pair representation:
\begin{equation}
h^{e}_{[CLS]} = BERT(E^1, E^2)
\end{equation}
The token-level discourse pair representation $h^d_{[CLS]}$ can be obtained in the same way for DRR. 

To project the token-level knowledge, we use the same BERT for event pair and discourse pair encoding.
During the optimization process, it is fine-tuned using the supervision signals from both ERE and DRR.

\subsection{Semantic Adaptor} 
\label{subsec:semantic_adaptor}

Because narrative and discourse analyses need to accurately represent the deeper semantic of the instance pairs, the shallow token-level knowledge captured by the BERT-based token encoder is not enough.
However, BERT always induces a non-smooth anisotropic semantic space which is adverse for semantic modelling of large-grained linguistic units~\citep{li2020sentence}.

To address this issue, we introduce an variational autoencoder-based (VAE-based) semantic encoder to represent the semantics of both events and clauses by transforming the anisotropic semantic distribution to a smooth and isotropic Gaussian distribution~\citep{kingma2014autoencoding,rezende2014stochastic,sohn2015learning}. 
To better learn the semantic encoder, the semantic adaptor shares the parameters $\theta^{Semantic}$ of it between ERE and DRR and train it using both classification supervision signals and KL divergence.

\begin{figure}[!t]
\centering
\includegraphics[width=0.5\columnwidth]{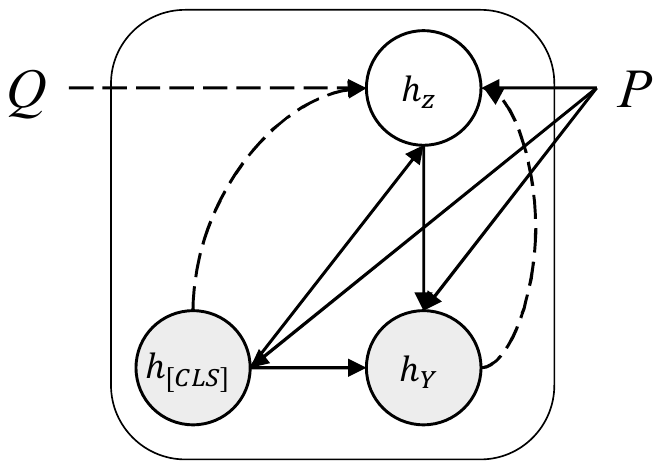}
\caption{The illustration of the semantic encoder as a directed graph.
We use solid lines to denote the generative model $P = p(h_Y|h_{[CLS]}, h_z)p(h_z|h_{[CLS]})$, and dashed lines to denote the variational approximation $Q = q(h_z|h_{[CLS]}, h_Y)$.
Both variational parameters and generative parameters are learned jointly.
}
\label{fig:VAE}
\end{figure}

Specifically, VAE is a directed graphical model with the generative model $P$ and the variational model $Q$, which learns the semantic representation $h_z$ of the input by an autoencoder framework.
Figure~\ref{fig:VAE} illustrates the graphic representation of the semantic encoder.
Specifically, we assume that there exists a continuous latent variable $h_z \sim \mathcal{N}(\mu, diag(\sigma^2))$, where $\mu$ and $\sigma^2$ are mean and variance of the Gaussian distribution respectively.
With this assumption, the original conditional probability of the event/discourse relations can be expressed by the following formula:
\begin{equation}
\begin{aligned}
p(h_Y|h_{[CLS]}) = \int_{h_z} &p(h_Y|h_{[CLS]}, h_z) \\
&p(h_z|h_{[CLS]}) d_{h_z}
\end{aligned}
\end{equation}

The posterior approximation is $q(h_z|h_{[CLS]}, h_Y)$, where $h_{[CLS]}$ can be $h^{e}_{[CLS]}$ or $h^d_{[CLS]}$ and $h_Y$ can be $h^e_Y$ or $h^d_Y$ according to the different tasks.
We 1) first obtain the input- and output-side representations via the shared BERT-based token encoder and the individual relation embedding networks, i.e., $h_{[CLS]}$ and $h_Y$;
2) then perform a non-linear transformation that project them onto the semantic space:
\begin{equation}
h_z' = tanh(W_z[h_{[CLS]}; h_Y]+b_z)
\end{equation}
3) obtain the above-mentioned Gaussian parameters $\mu$ and $log \sigma^2$ through linear regression:
\begin{equation}
\mu = W_{\mu}h_z' + b_{\mu}, \ \ \ log{\sigma}^2 = W_{\sigma}h_z' + b_{\sigma}
\end{equation}
where $W$ and $b$ are the parameter matrix and bias term respectively;
4) use a reparameterization trick~\citep{kingma2014autoencoding,sohn2015learning} to get the final semantic representation:
\begin{equation}
h_z = \mu + \sigma \odot \epsilon
\end{equation}
where $\epsilon \sim \mathcal{N}(0, I)$ and $h_z$ can be $h^e_z$ or $h^d_z$.

The neural model for the prior $p(h_z|h_{[CLS]})$ is the same as that for the posterior $q(h_z|h_{[CLS]}, h_Y)$,  except for the absence of $h_Y$.
Besides, those two models have parameters independent of each other.

During testing, due to the absence of the output-side representation $h_Y$, we set $h_z$ to be the mean of $p(h_z|h_{[CLS]})$~\citep{zhang2016variationalb}, i.e., $\mu$.
During training, we minimize the Kullback-Leibler divergence $KL(P||Q)$ between the generation model $P$ and the inference model $Q$.
Intuitively, KL divergence connects these two models:
\begin{equation}\label{eqa:kl_divergence}
KL(q(h_z|h_{[CLS]}, h_Y)||p(h_z|h_{[CLS]})) 
\end{equation}

To project the semantic-level knowledge, we use the same VAE for both event pair and discourse pair.
Therefore, the commonalities of event semantics and discourse semantics can be captured more accurately.

\subsection{Coarse Category Adaptor} 
\label{subsec:coarse_category_adaptor}

The token adaptor and the semantic adaptor commendably cover the knowledge entailed on the input-side.
In addition, we found that ERE and DRR share the same coarse-grained categories: \emph{Temporal}, \emph{Contingency}, \emph{Comparison} and \emph{Expansion}~\citep{prasad2008penn,zhang2020aser}, although they have different fine-grained categories. 

To this end, we design the coarse category adaptor in a coarse-to-fine framework~\citep{petrov2009coarse} to bridge the gap between the heterogeneous fine-grained targets.
Specifically, we share the parameters $\theta^{Coarse}$ of the coarse-grained classifier and the coarse label embedding network to obtain more credible coarse category representations.

Specifically, we first use the token representation $h_{[CLS]}$ and the semantic representation $h_z$ to predict the coarse-grained labels: 
\begin{equation}
Y^{c} = Classifier^{Coarse}(h_{[CLS]}, h_z)
\end{equation}
where $Y^c \in$ \{\emph{Temporal}, \emph{Contingency}, \emph{Comparison}, \emph{Expansion}\}.
After that, we use the coarse label embedding network to obtain the corresponding coarse-grained label embedding $h_{Y^c}$, which is referred as the coarse category representation.

To project that label-level knowledge, we use the same coarse-grained classifier and the same coarse label embedding network.
During the optimization process, both event instances and discourse instances can be used to train this coarse category encoder.
The more supervision signals make it more effective.

\subsection{Full Model Training} 
\label{subsec:full_model_training}

In this paper, we utilize multi-task learning~\citep{caruana1997multitask} to implement the knowledge projection from discourse to narrative.
It expects correlative tasks (ERE and DRR) can help each other to learn better by sharing the parameters of three adaptors.
Given ERE and DRR training datasets, an alternate optimization approach~\citep{dong2015multi} is used to optimizate MKPNet:
\begin{equation}
\begin{aligned}
L(\theta) = &\alpha (L(\theta; Y) + \lambda KL(P||Q)) \\
&+ (1 - \alpha)  L(\theta; Y^{c})
\end{aligned}
\end{equation}
where $Y$ can be $Y^{im}$ or $Y^d$ according to the different tasks, $\lambda$, $\alpha$ are two hyperparameters, $KL(P||Q))$ is the KL divergence in the semantic encoder, $L(\theta; Y)$ and $L(\theta; Y^c)$ are fine-grained and coarse-grained objectives respectively:
\begin{align}
&L(\theta; Y) = \log p(Y|h_{[CLS]}, h_z, h_{Y^c}) \\
&L(\theta; Y^c) = \log p(Y^c|h_{[CLS]}, h_z)
\end{align}
It should be noticed that in MKPNet, \{$\theta^{BERT}, \theta^{Semantic}, \theta^{Coarse}$\} are the shared parameters of the BERT-based token encoder, the VAE-based semantic encoder and the coarse category encoder between ERE and DRR. 
And  \{$\theta^{Fine}$\} are separated parameters of the fine-grained ERE and DRR classifiers.

\section{Experiments} 
\label{sec:experiments}

We conduct intrinsic experiments on ASER~\citep{zhang2020aser} to assess the effectiveness of the proposed MKPNet, and extrinsic experiments on WSC~\citep{levesque2012winograd} to verify the value of the extracted event relations.

\subsection{Intrinsic Experiments}

\textbf{Datasets.} 
For discourse relation recognition (DRR), we use PDTB 2.0~\citep{prasad2008penn} with the same splits of~\citet{ji2015one}: sections 2-20/0-1/21-22 respectively for train/dev/test. 
For event relation extraction (ERE), because there is no labelled training corpus, we construct a new dataset by removing the connectives of the explicit event relation instances in ASER core version\footnote{
\url{https://hkust-knowcomp.github.io/ASER}} and retaining at most 2200 instances with the highest confidence scores for each category\footnote{
Higher confidence score means more credible instance.}.  
In this way, we obtain 23,181/1400/1400 train/dev/test instances – we denoted it as implicit event relation extraction (IERE) dataset.

\textbf{Implementation.} 
We implement our model based on pytorch-transformers~\citep{wolf2020transformers}. 
We use BERT-base and set all hyper-parameters using the default settings of the SotADRR model~\citep{kishimoto2020adapting}.

\textbf{Baselines.}
For ERE, we compare the proposed MKPNet with the following baselines:
\begin{itemize}
\item Baselines w/o Discourse Knowledge are only trained on IERE training set. 
We choose the BERT-CLS as the representative of them due to its SotAperformance.
\item Baselines with Discourse Knowledge improve the learning of ERE via transfer learning~\citep{pan2009survey,pan2010domain} from discourse models, i.e., first pre-train a parameter prior on PDTB 2.0 and then fine-tune it on IERE --– we denote it as BERT-Transfer.
\end{itemize}

For DRR, we compare the proposed MKPNet with the following baselines:
\begin{itemize}
\item \citet{bai2018deep} is a deep neural network model augmented by variable grained text representations like character, sentence and sentence pair levels.
\item \citet{kishimoto2020adapting} is the SotADRR model, BERT-CLS, which incorporating BERT with one additional output layer.
\end{itemize}

\subsubsection{Overall Results} 
\label{subsubsec:overall_results}

Table~\ref{tab:num_relations}-\ref{tab:IDRR_results} show the overall ERE/DRR results of baselines and MKPNet.
For our approach, we use the full MKPNet and its four ablated settings: 
MKPNet w/o SA, MKPNet w/o CA, MKPNet w/o SA \& CA and MKPNet w/o KP, where SA, CA and KP denote semantic adaptor, coarse category adaptor and knowledge projection correspondingly.
We can see that:

\begin{table}[!t]
\centering
\resizebox{0.95\columnwidth}{!}{
\begin{tabular}{lc}  
\hline 
& \textbf{Number of Relations} \\
\hline 
\hline 
FrameNet~\citep{baker1998berkeley} & 1,709  \\
ConceptNet~\citep{speer2017conceptnet} & 116,097 \\
Event2Mind~\citep{smith2018event2mind} & 57,097 \\
ATOMIC~\citep{sap2019atomic} & 877,108 \\
Knowlywood~\citep{tandon2015knowlywood} & 2,644,415 \\
ASER~\citep{zhang2020aser} & 1,287,059 \\
\hline 
ASER++ (core) & 2,034,963 \\
ASER++ (high) & 3,530,771 \\
ASER++ (full) & 8,766,098 \\
\hline 
\end{tabular}
}
\caption{Number comparison of event relations in ASER++ and existing event-related resources.}
\label{tab:num_relations}
\end{table}

\begin{table}[!t]
\centering
\resizebox{0.95\columnwidth}{!}{
\begin{tabular}{lll}  
\hline 
& \ \textbf{Acc} & \ \textbf{F1} \\
\hline 
\hline 
\multicolumn{3}{l}{\textbf{Baselines w/o Discourse Knowledge}} \\
\quad BERT-CLS & 53.00 & 52.24 \\
\quad MKPNet w/o KP & 53.94 & 53.52 \\
\hline 
\multicolumn{3}{l}{\textbf{Baselines with Discourse Knowledge}} \\
\quad BERT-Transfer & 54.29 & 53.44 \\
\hline 
\multicolumn{3}{l}{\textbf{Multi-tier Knowledge Projection}} \\
\quad MKPNet w/o SA \& CA & 54.79$_{\triangle 0.85}$ & 53.90$_{\triangle 0.38}$ \\
\quad MKPNet w/o CA & 55.14$_{\triangle 1.20}$ & 54.42$_{\triangle 0.90}$ \\
\quad MKPNet w/o SA & 55.29$_{\triangle 1.35}$ & 54.92$_{\triangle 1.40}$ \\

\quad MKPNet & \textbf{55.86$_{\triangle 1.92}$} & \textbf{55.36$_{\triangle 1.84}$} \\
\hline 
\end{tabular}
}
\caption{Experimental results on IERE test set, where $_{\triangle}$ means the improvements when compared with MKPNet w/o KP.
All improvements of MKPNet are statistical significance at p\textless0.01 over the baseline MKPNet w/o KP.}
\label{tab:overall_results}
\end{table}

\begin{table}[!t]
\centering
\resizebox{0.8\columnwidth}{!}{
\begin{tabular}{lc}  
\hline 
Model & \textbf{Acc}\\
\hline 
\hline 
\citet{bai2018deep} & 48.22 \\
BERT-CLS~\citep{kishimoto2020adapting} & 51.40 \\
BERT-CLS (Ours) & 50.91 \\
\hline 
MKPNet w/o KP & 52.86 \\
MKPNet & \textbf{54.09} \\
\hline 
\end{tabular}
}
\caption{Experimental results on PDTB 2.0 test set.
For a fair comparison, the results of baselines are adapted from their original papers.}
\label{tab:IDRR_results}
\end{table}

\textbf{1. Based on MKPNet, we enrich the original ASER by abundant implicit event relations.}
Considering the computational complexity, we classify the event pairs co-occurrence in the same document and filter them by confidence scores.
Specifically, we compute the confidence score by multiplying the classification probability and the frequency of the event pair.
Integrating with the original explicit event relations, we can obtain the enriched EventKGs ASER++ (core/high/full) with the different threshold confidences (3/2/1).
Table~\ref{tab:num_relations} shows that when compared with existing event-related resources, ASER++ has an overwhelming advantage in the number of event relations.

\textbf{2. The proposed MKPNet achieves SotAperformance for ERE.}
MKPNet can significantly outperform the BERT-Transfer and achieves 55.86 accuracy and 55.36 F1. 
MKPNet w/o KP obtains considerable performance improvements when compared with BERT-CLS.
We believe this is because MKPNet fully explores the knowledge on different tiers, and modelling knowledge tier-by-tier is effective.

\textbf{3. By projecting knowledge at token-level, semantic level and label level, all three adaptors are useful and are complementary with each other.}
When compared with the full model MKPNet, its four variants show declined performance in different degrees.
MKPNet outperforms MKPNet w/o CA 0.72 accuracy and 0.94 F1, which indicates that our coarse category adaptor successfully bridges the gap of heterogeneous fine-grained targets.
MKPNet outperforms MKPNet w/o SA 0.57 accuracy and 0.44 F1, and therefore we believe that our latent semantic adaptor is helpful for capture the semantic-level commonalities.
Finally, there is a significant decline between MKPNet w/o KP and MKPNet w/o SA \& CA, which means that token adaptor is indispensable.
The insight in those observations is that the commonalities between discourses and narratives under the hierarchical structure, thus projecting them at different levels is effective, and three adaptors can be complementary with each other.

\textbf{4. The commonalities between discourses and narratives are beneficial for both ERE and DRR.}
Compared with the baselines w/o discourse knowledge --- BERT-CLS and MKPNet w/o KP, both the naive transfer method --- BERT-Transfer and our MKPNet achieve significant performance improvements:  
BERT-Transfer gains 1.29 accuracy and 1.20 F1 when compared to BERT-CLS, and MKPNet gains 1.92 accuracy and 1.84 F1 when compared to MKPNet w/o KP.
Besides, for DRR, our method MKPNet also substantially outperforms the other baselines and its variant MKPNet w/o KP.
These results verified the commonalities between discourse knowledge and narrative knowledge.

\subsubsection{Detailed Analysis} 
\label{subsubsec:detailed_analysis}

\textbf{Effects of Semantic-level Knowledge and Label-level Knowledge.}
In these experiments, we compare the performance of our models, MKPNet, MKPNet w/o CA and MKPNet w/o SA with or without knowledge projection to find out the effects of semantic-level knowledge and label-level knowledge.
From Table~\ref{tab:sa_ca_adaptors}, we can see that:
(1) Compared with their counterparts, MKPNet, MKPNet w/o CA and MKPNet w/o SA with knowledge projection lead to significant improvements.
Thus, it is convincing that the performance improvements mainly come from the discourse knowledge rather than the neural architecture;
(2) Current knowledge projection can be further improved by exploiting more accurate discourse knowledge: MKPNet w/o SA*, which uses golden coarse categories, achieves striking performance (Acc 70.50; F1 70.32).

\begin{figure}[!t]
\centering
\includegraphics[width=0.95\columnwidth]{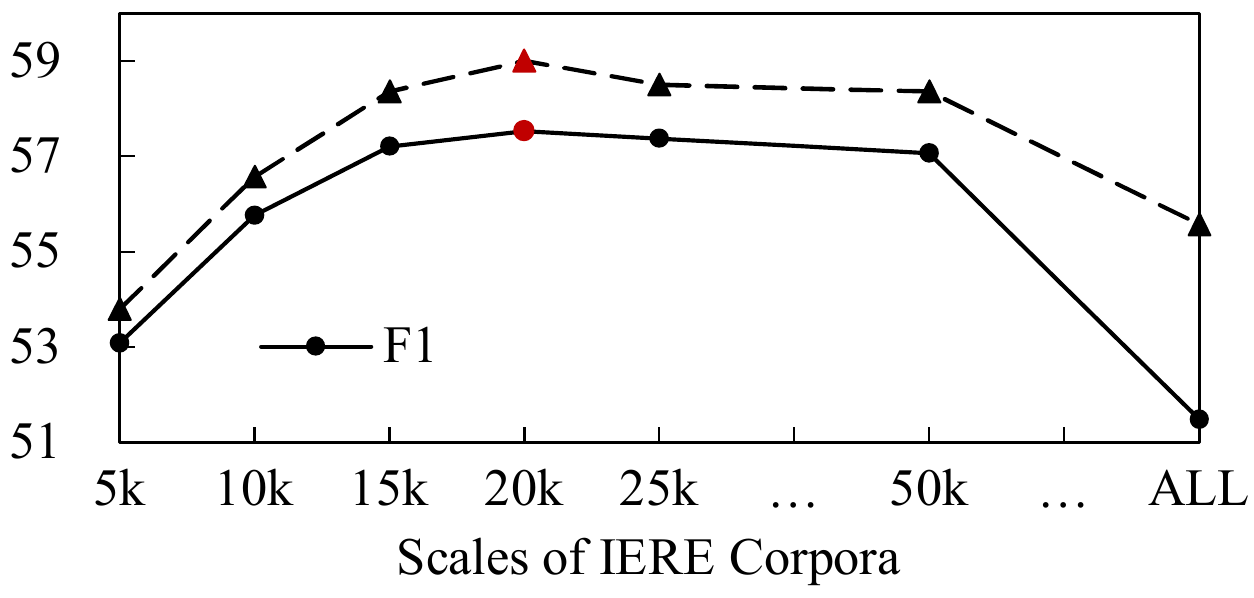}
\caption{Experimental results of using different sizes of IERE training corpora.
}
\label{fig:size_results}
\end{figure}

\textbf{Tradeoff between Dataset Quality and Size.}
As described above, the IERE training dataset is constructed using the most confident instances in ASER core version.  
We can construct a larger but lower quality dataset by incorporating more instances with lower confidence, i.e., the quality-size tradeoff problem. 
To analyze the tradeoff between the quality and size, we construct a set of datasets with different sizes/qualities, and Figure~\ref{fig:size_results} shows the corresponding results of MKPNet on the development set. 
We can see that the size is the main factors for performance improvements at the beginning: 
every 5,000 additional instances can result in a significant improvement (about 2 to 3 F1 gain). 
When the size is large (more than 20,000 instances in our experiments), more instances will not result in performance improvements, and the low-quality instances will hurt the performance. 

\begin{table}[!t]
\centering
\resizebox{0.95\columnwidth}{!}{
\begin{tabular}{l|c|cc|cc}  
\hline 
& \multirow{2}*{\textbf{KP}} & \multicolumn{2}{c|}{\textbf{Fine-grained}} & \multicolumn{2}{c}{\textbf{Coarse-grained}} \\
\cline{3-6}
& & \textbf{Acc} & \textbf{F1} & \textbf{Acc} & \textbf{F1} \\
\hline 
\hline 
BERT-CLS & & 53.00 & 52.24 & --- & --- \\
\hline 
\hline 
\multirow{2}*{MKPNet} & & 53.94 & 53.52 & --- & --- \\
& \checkmark & \textbf{55.86} & \textbf{55.36} & --- & --- \\
\hline 
\hline 
\multirow{2}*{MKPNet w/o CA} & & 53.79 & 53.39 & --- & --- \\
& \checkmark & \textbf{55.14} & \textbf{54.42} & --- & --- \\
\hline 
\hline 
\multirow{2}*{MKPNet w/o SA} & & 53.21 & 52.48 & 66.57 & 63.04 \\
& \checkmark & \textbf{55.29} & \textbf{54.92} & \textbf{67.93} & \textbf{64.76} \\
\hline 
MKPNet w/o SA\textbf{*} & & 70.50 & 70.32 & 100.0 & 100.0 \\
\hline 
\end{tabular}
}
\caption{Effect of semantic-level knowledge and label-level knowledge on IERE test set, where \textbf{KP} stands for knowledge projection and \textbf{*} stands for golden coarse-grained categories.}
\label{tab:sa_ca_adaptors}
\end{table}

\subsection{Extrinsic Experiments}

The above intrinsic experiments verified the effectiveness of the proposed MKPNet for ERE. 
In this section, we use the core version of our enriched EventKGs --- ASER++, and then conduct extrinsic experiments on Winograd Schema Challenge (WSC)~\citep{levesque2012winograd} to verify the effect of ASET++.

\begin{figure}[!t]
\centering
\includegraphics[width=0.95\columnwidth]{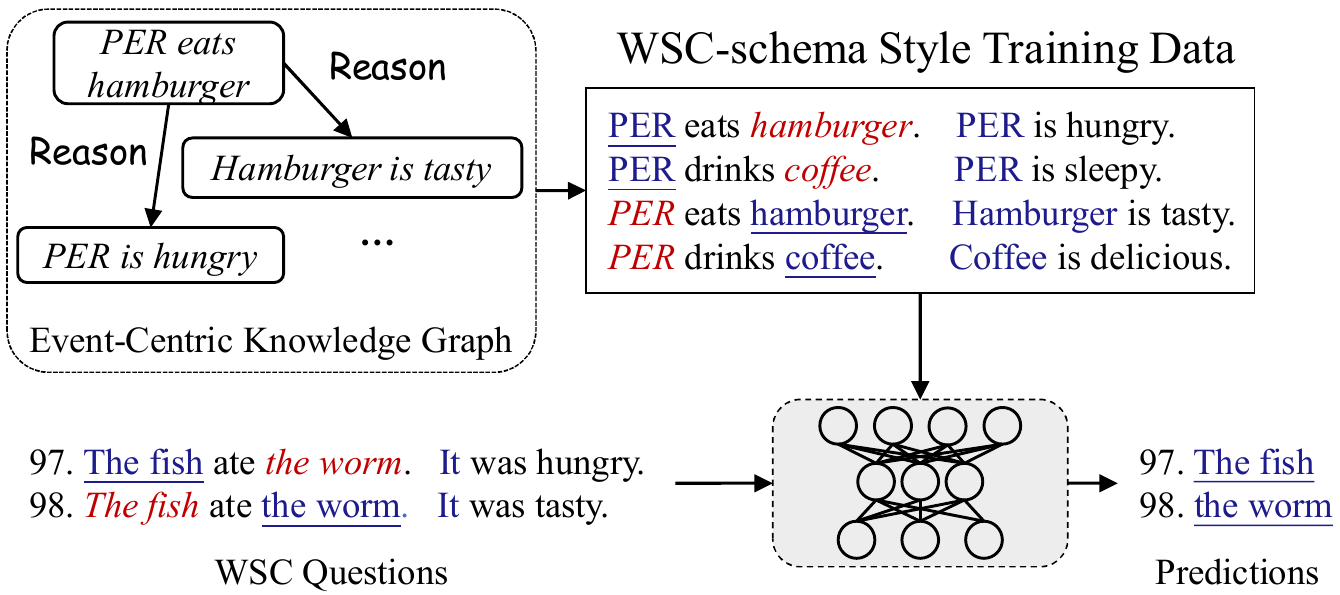}
\caption{An overview of WSC implementation.
The blue color means the correct reference while the red color means the wrong one.
}
\label{fig:wsc_example}
\end{figure}

\textbf{WSC Implementation.}
WSC is challenging since its schema is a pair of sentences that differ only in one or two words and that contain a referential ambiguity that is resolved in opposite directions in the two sentences.
According to ~\citet{kocijan2019surprisingly}, fine-tuning pre-trained language models on WSC-schema style training sets is a robust method to tackle WSC.
Therefore, as Figure~\ref{fig:wsc_example} shows, we transform ASER++ to WSC-schema style training data in the same way as~\citet{zhang2020aser} and fine-tune BERT on it, which we refer to as BERT (ASER++).
We compare BERT (ASER++) with these baselines:
\begin{itemize}
\item \textbf{Pure Knowledge-based Methods} are heuristical rule-based methods, such as Knowledge Hunting~\citep{emami2018knowledge} and String Match~\citep{zhang2020aser}.
\item \textbf{Language Model-based Methods} use language model trained on large-scale corpus and tuned specifically for the WSC task, such as LM~\citep{trinh2018simple}.
\item \textbf{External Knowledge Enhanced Methods} are models based on BERT and trained with the different external knowledge resource, e.g., WscR~\citep{rahman2012resolving,kocijan2019surprisingly}
\end{itemize}

We implement our model based on pytorch-transformers~\citep{wolf2020transformers}.
BERT-large is used.
All hyper-parameters are default settings as~\citet{kocijan2019surprisingly}.

\begin{table}[!t]
\centering
\resizebox{0.95\columnwidth}{!}{
\begin{tabular}{lc}  
\hline 
Model & WSC\\
\hline 
\hline 
\multicolumn{2}{l}{\textbf{Pure Knowledge-based Methods}} \\
\quad Knowledge Hunting~\citep{emami2018knowledge} & 57.3 \\
\quad String Match~\citep{zhang2020aser} & 56.6 \\
\hline
\multicolumn{2}{l}{\textbf{Language Model-based Methods}} \\
\quad LM (Single)~\citep{trinh2018simple} & 54.5 \\
\quad LM (Ensemble)~\citep{trinh2018simple} & 61.5 \\
\quad BERT (w/o finetuning)~\citep{devlin2019bert}& 61.9 \\
\hline
\multicolumn{2}{l}{\textbf{External Knowledge Enhanced Methods}} \\
\quad BERT (WscR)~\citet{kocijan2019surprisingly} & 71.4 \\
\quad BERT (ASER)~\citet{zhang2020aser} & 64.5 \\
\quad BERT (ASER \& WscR)~\citet{zhang2020aser} & 72.5 \\
\quad BERT (ASER++) & 66.2 \\
\quad BERT (ASER++ \& WscR) & \textbf{74.1} \\
\hline
\end{tabular}
}
\caption{The overall results of extrinsic experiments.
The evaluation metric is accuracy.}
\label{tab:overall_nlu}
\end{table}

\textbf{Extrinsic Results.}
Table~\ref{tab:overall_nlu} shows the overall results of extrinsic experiments.
We can see that: 
By fine-tuning BERT on our enriched EventKG --- ASER++, the WSC performance can be significantly improved. 
BERT (ASER++) and BERT (ASER++ \& WscR) outperform BERT (ASER) and BERT (ASER \& WscR) respectively, which verified the effectiveness of ASER++ and implicit event relations are beneficial for downstream NLU tasks.

\section{Related Work}

\textbf{Event-centric Knowledge Graphs.}
Knowledge graphs have come from entity-centric ones~\citep{banko2007open,suchanek2007yago,bollacker2008freebase,wu2012probase} to event-centric ones.
However, the construction of traditional KGs takes domain experts much effort and time, which are often with limited size and cannot effectively resolve real-world applications, e.g., FrameNet~\citep{baker1998berkeley}.
Recently, many modern and large-scale KGs have been built semi-automatically, which focus on events~\citep{tandon2015knowlywood,rospocher2016building,gottschalk2018eventkg,zhang2020aser} and commonsense~\citep{speer2017conceptnet,smith2018event2mind,dalviand2018tracking,sap2019atomic}.
Specifically, \citet{yu2020enriching} proposes an approach to extract entailment relations between eventualities, e.g., \emph{``I eat an apple''} entails \emph{``I eat fruit''}, and release an event entailment graph (EEG).
Different from EEG, this paper focuses on implicit event relations which are not extracted due to the absences of the connectives and discontinuity.

\textbf{Knowledge Transfer.} 
Due to the data scarcity problem, many knowledge transfer studies have been proposed, including multi-task learning~\citep{caruana1997multitask}, transfer learning~\citep{pan2009survey,pan2010domain}, and knowledge distillation~\cite{hinton2014distilling}.
Recently, researchers are interested in training/sharing/transferring/distilling models layer by layer to fully excavate the knowledge~\cite{pan2016domain,guo2019spottune,kang2020decoupling,li2020bert}.
In this paper, we propose a knowledge projection method which can project discourse knowledge to narraties on different tiers.

\section{Conclusions}
In this paper, we propose a knowledge projection paradigm for event relation extraction and \textbf{M}ulti-tier \textbf{K}nowledge \textbf{P}rojection \textbf{N}etwork (\textbf{MKPNet}) is designed to leverage multi-tier discourse knowledge. 
By effectively projecting knowledge from discourses to narratives, MKPNet achieves the new state-of-the-art event relation extraction performance, and extrinsic experimental results verify the value of the extracted event relations.  
For future work, we want to design new data-efficient algorithms to learn effective models using low-quality and heterogeneous knowledge.

\section*{Acknowledgments}
This work is supported by the Strategic Priority Research Program of Chinese Academy of Sciences, Grant No. XDA27020200, the National Natural Science Foundation of China under Grants no. U1936207, and in part by the Youth Innovation Promotion Association CAS (2018141).

\bibliographystyle{acl_natbib}
\bibliography{acl2021}

\end{document}